# Prompting Science Report 1: Prompt Engineering is Complicated and Contingent

Lennart Meincke[1,2], Ethan Mollick[1], Lilach Mollick[1], Dan Shapiro[1,3]

[1] Generative AI Labs, The Wharton School of Business, University of Pennsylvania
[2] WHU–Otto Beisheim School of Management,
[3] Glowforge

## Summary

This is the first of a series of short reports that seek to help business, education, and policy leaders understand the technical details of working with AI through rigorous testing. In this report, we demonstrate two things:

- There is no single standard for measuring whether a Large Language Model (LLM) passes a benchmark, and that choosing a standard has a big impact on how well the LLM does on that benchmark. The standard you choose will depend on your goals for using an LLM in a particular case.
- It is hard to know in advance whether a particular prompting approach will help or harm the LLM's ability to answer any particular question. Specifically, we find that sometimes *being polite* to the LLM helps performance, and sometimes it lowers performance. We also find that constraining the AI's answers helps performance in some cases, though it may lower performance in other cases.

Taken together, this suggests that benchmarking AI performance is not one-size-fits-all, and also that particular prompting formulas or approaches, like being polite to the AI, are not universally valuable.

## How we Benchmark the AI

In this Report, we are not particularly interested in providing results for any one model or benchmark, but rather to show that, given a model and a benchmark, small changes can lead to big variations. To do so, we pick two of the most popular models at the time of testing: GPT-4o (gpt-4o-2024-08-06) and GPT-4o-mini (gpt-4o-mini-2024-07-18). Neither GPT-4o nor 4o-mini are the most advanced available, nor do they use reasoning methods. We would expect that more advanced models would score significantly higher in benchmarks. However, the goal of this Report is to study variability within models, not model differences.

For a benchmark, we selected the commonly-used GPQA Diamond (Graduate-Level Google-Proof Q&A Benchmark) dataset (Rein et al. 2024). The GPQA Diamond set comprises 198 multiple-choice PhD-level questions across biology, physics, and chemistry. This is a challenging test: PhDs in the corresponding domains reach 65% accuracy (74% when discounting clear mistakes the experts identified in retrospect), while highly skilled non-expert validators only reach 34% accuracy, despite spending on average over 30 minutes with unrestricted access to the web (i.e., the questions are "Google-proof)" (Rein et al. 2024).

Many benchmarking attempts ask the AI to answer each question a single time, but AI results can often vary, even when asked the same question. To be more rigorous, we ask each question 100 times for each prompting condition, thus providing a deeper insight into the consistency and reliability of the models' responses (Miller 2024). Since we have many attempts for each question, we need to make decisions about what constitutes a "correct" answer. Interestingly, different AI labs use different standards at different times, so we wanted to establish three clear ways of "passing" and why they might be useful.

- **Complete Accuracy/100% Correct**: This condition requires the AI to get the correct answer 100% of the time, without failing on any of the 100 attempts. This standard will be most appropriate for situations where no errors can be tolerated.
- **High Accuracy/90% Correct:** In this condition, the AI must give the right answer 90% of the time, failing on no more than 10 out of 100 attempts. A similar standard might be appropriate where human-level error is tolerated.
- **Majority Correct/51% Correct:** In this condition the AI needs to get the right answer in the majority of the 100 attempts, failing no more than 49 out of 100 attempts. A similar standard might be appropriate when the AI is consulted multiple times, and the majority answer is selected.

These standards are much more rigorous than two of the most common standards used in computer science for AI evaluation:

- **PASS@100:** In the PASS@100 standard, one right answer out of 100 attempts would be considered correct, meaning the answer can be wrong 99 times out of 100. The PASS standard is most often done with smaller tests, like PASS@5 (1 right out of 5 attempts) or even PASS@1, which would be the equivalent of getting the right answer on a single test.
- **CONSENSUS@100**: This is different from our consensus measurement above, as it selects the modal answer. Thus getting 26 answers right out of 100 (assuming four possible answers to select from) would be a success.

While those approaches are useful for the development of AI systems, they are not generally as appropriate for benchmarking for real-world applications.

## How we Prompt the AI

We use the standard approach for the GPQA, the zero-shot reference implementation by Rein et al. (2024). The temperature for each request is 0. Since there are four questions, random guessing would result in 25% correct answers. Here is an example of a question, and where we insert any prefixes or suffixes (following the reference implementation):

> *[prefix] Two quantum states with energies E1 and E2 have a lifetime of 10^-9 sec and 10^-8 sec, respectively. We want to clearly distinguish these two energy levels. Which*

*one of the following options could be their energy difference so that they can be clearly resolved?*

*Choices:*
*(A) 10^-9 eV*
*(B) 10^-11 eV*
*(C) 10^-8 eV*
*(D) 10^-4 eV*

*[suffix]*

- **Baseline (formatted) prompt**  The reference approach adds a prefix ("What is the correct answer to this question") and suffix ("Format your response as follows: 'The correct answer is (insert answer here)'"). It further uses a system prompt ("You are a very intelligent assistant, who follows instructions directly.") for each request[1].
- **Unformatted prompt:** In this variation, we remove the suffix from the baseline that told the AI to format its responses in a particular way. This mimics the more natural way that people ask questions of the AI, and previous research suggests that formatting may limit AI performance (Tam et al. 2024).
- **Polite prompt:** We change the prefix to "Please answer the following question." Whether or not being polite to LLMs changes the result has been an ongoing question in both practice and research (Yin et al. 2024).
- **Commanding prompt:** We change the prefix to "I order you to answer the following question." We selected this as a "less polite" contrast to the Polite Prompt.

Each prompt condition was tested 100 times each across all 198 questions of the Diamond GPQA dataset (19,800 runs per prompt per model).

## Results

We find substantial performance variability across measurements, indicating that many questions are not answered correctly consistently (Figure 1, Supplementary Table 1). Notably, using the formatted prompt and 100% correct condition, both GPT-4o and GPT-4o mini only perform 5 percentage points (RD = 0.051; 95% CI [-0.035, 0.136]; p = 0.267) and 4.5 percentage points (RD = 0.045; 95% CI [-0.040, 0.136]; p = 0.345) respectively better than a random guess on a PhD-level benchmark, both differences insignificant. At 90% correct answers, 4o performs significantly better than a random guess (RD = 0.111; 95% CI [0.020, 0.197]; p = 0.016), whereas 4o-mini only does so at the 51% threshold (RD = 0.141; 95% CI [0.051, 0.237]; p = 0.003). Overall, 4o beats a random guess in 5 out of 12 comparisons at varying thresholds (once at 90%, 4 times at 51%), and 4o-mini beats a random guess in 4 (all at 51%) out of 12 comparisons (see Supplementary Table 2 for all comparisons). While GPT-4o achieves better results than GPT-4o mini in the formatted condition, the results are not significantly different at majority (RD = 0.066; 95% CI [-0.015, 0.146]; p = 0.113), 90% correct (RD = 0.030; 95% CI [-0.051, 0.111]; p = 0.461) or 100% correct (RD = 0.005; 95% CI [-0.071, 0.081]; p = 0.884). Unsurprisingly, the majority performance is significantly better than 100% correct for GPT-4o (RD = -0.177; 95% CI [-0.232, -0.116]; p < 0.001) and GPT-4o mini (RD = -0.106; 95% CI [-0.152, -0.061]; p < 0.001) in the formatted condition.

---
[1] We find no evidence that removing the system prompt noticeably affects average performance, but question-level performance may vary.

In our case, we find that using the unformatted prompt, model performance drops significantly for GPT-4o (RD = 0.086; 95% CI [0.040, 0.136]; p < 0.001) and GPT-4o mini (RD = 0.121; 95% CI [0.056, 0.187]; p < 0.001). We find no evidence that suggests that basic prompt engineering produces significant effects between the three conditions, except for 4o mini "I order" vs. "Please" at the 51% threshold (see Supplementary Table 1).

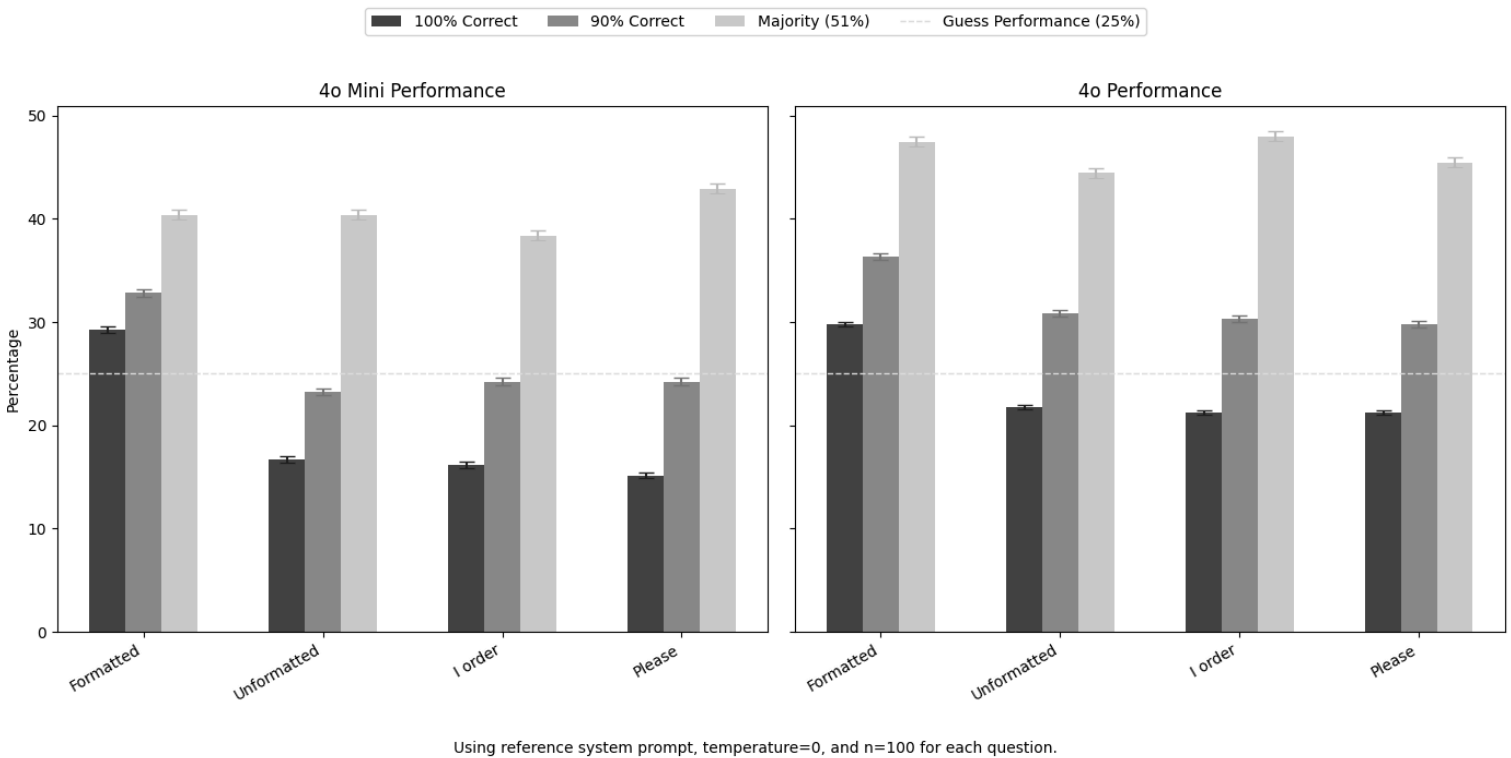

**Figure 1:** GPT 4o mini and 4o Performance across conditions. Error bars show 95% confidence intervals for individual proportions. For statistical comparisons between conditions, see Supplementary Table 1.

Since we test each question 100 times, we can look at whether the Polite and Commanding prompts help at the level of individual questions. Interestingly, at the question-level, we can find significant differences between many questions (see Figure 2). Supplementary Table 3 contains the full comparisons. These differences disappear when aggregating across all questions as seen above, but suggest that specific prompting techniques might work for specific questions for unclear reasons.

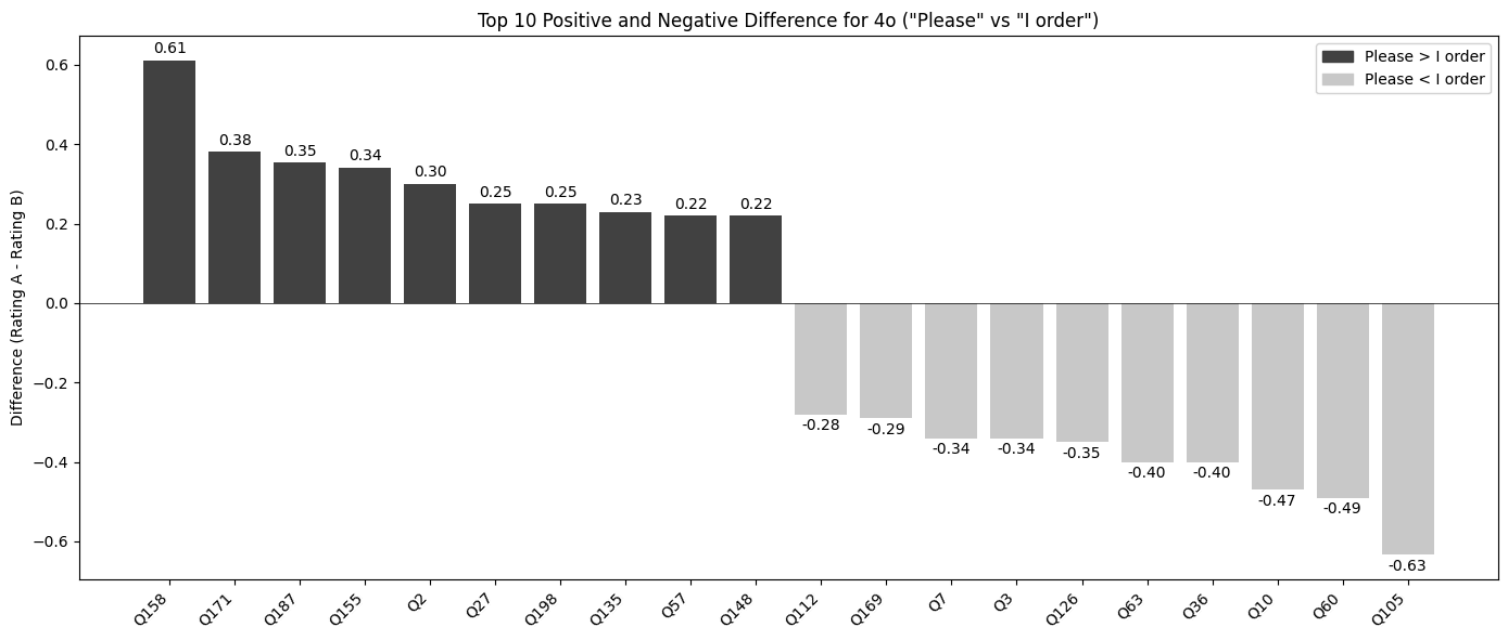

**Figure 2:** Top-10 performance differences for GPT-4o in the "Please" and "I order" conditions. All differences are highly significant (p < 0.01) and uncorrected. Supplementary Table 3 contains confidence intervals and statistics.

## Discussion

There are a few important take-aways from these results.

- LLMs can be inconsistent when answering questions. By looking at 100 attempts, not just one, we find that other benchmarking efforts can substantially overestimate model reliability, which is highly variable. This was on a difficult PhD-level benchmark, the GPQA Diamond, and may not apply to all benchmarks, nor to larger models, but deserves careful consideration when deciding when to use an LLM.
- The standard for benchmarking is important. At higher correctness thresholds, neither GPT-4o nor GPT-4o mini significantly outperformed random guessing, while at lower thresholds they did. Future work should justify the standards used for measuring AI performance.
- Prompting "tricks" like politeness are not universal. While we observed significant differences in performance across individual questions under all conditions, such differences became negligible once aggregated across the entire dataset. This indicates that prompt variations, although impactful at the question level, are dwarfed by overall model characteristics.
- Formatting is consistently important. Removing explicit formatting constraints consistently led to performance degradation for both GPT-4o variants. These findings align with existing literature (Salido et al. 2025) that underscores the susceptibility of

LLM benchmark performance to minor changes. However, whether or not formatting helps or hurts in a particular setting is likely to vary by model and setting.

These results show that methodological rigor, and particularly repeated sampling, provides a clearer picture of genuine model knowledge and consistency, highlighting the crucial role measurement methods play in evaluating LLM capabilities.

## Conclusion

Our results demonstrate that how we measure performance greatly influences our interpretations of LLM capabilities. Traditional reporting methods reliant on single or mode-based correctness metrics can mask significant inconsistencies in model performance. The current study's adoption of repeated sampling reveals profound variability at higher correctness thresholds. Moreover, prompt variations, although impactful at the question-level, become secondary when viewed through rigorous, repeated measures across an entire dataset.

## Supplemental Material

**Supplementary Table 1.** Pairwise-comparison results across all conditions for each model using paired bootstrap-permutation tests (5,000 replicates). P-values represent the proportion of permuted differences with absolute values exceeding observed differences under the null hypothesis. Questions were considered 'perfect' at ≥100 successes, with sensitivity analyses at thresholds of 90 and 51.

| Model | Comparison | Threshold % | RD [95% CI] | Statistics |
|---|---|---|---|---|
| 4o mini | Formatted vs Unformatted | 100 | 0.121 [0.056, 0.182] | p<0.001 |
| 4o mini | Formatted vs I order | 100 | 0.131 [0.071, 0.192] | p<0.001 |
| 4o mini | Formatted vs Please | 100 | 0.146 [0.081, 0.212] | p<0.001 |
| 4o mini | Unformatted vs I order | 100 | 0.005 [-0.020, 0.030] | p=0.799 |
| 4o mini | Unformatted vs Please | 100 | 0.015 [-0.015, 0.045] | p=0.415 |
| 4o mini | I order vs Please | 100 | 0.015 [-0.010, 0.040] | p=0.316 |
| 4o mini | Formatted vs Unformatted | 90 | 0.101 [0.040, 0.162] | p=0.001 |
| 4o mini | Formatted vs I order | 90 | 0.086 [0.025, 0.146] | p=0.007 |
| 4o mini | Formatted vs Please | 90 | 0.091 [0.035, 0.146] | p=0.003 |
| 4o mini | Unformatted vs I order | 90 | 0.000 [-0.040, 0.040] | p=1.000 |
| 4o mini | Unformatted vs Please | 90 | -0.010 [-0.051, 0.030] | p=0.678 |
| 4o mini | I order vs Please | 90 | 0.005 [-0.025, 0.035] | p=0.809 |
| 4o mini | Formatted vs Unformatted | 51 | -0.015 [-0.076, 0.040] | p=0.598 |
| 4o mini | Formatted vs I order | 51 | 0.010 [-0.045, 0.071] | p=0.781 |
| 4o mini | Formatted vs Please | 51 | -0.030 [-0.091, 0.035] | p=0.363 |
| 4o mini | Unformatted vs I order | 51 | 0.035 [-0.010, 0.081] | p=0.149 |
| 4o mini | Unformatted vs Please | 51 | 0.015 [-0.030, 0.061] | p=0.578 |
| 4o mini | I order vs Please | 51 | -0.066 [-0.111, -0.020] | p=0.006 |
| 4o | Formatted vs Unformatted | 100 | 0.086 [0.040, 0.136] | p<0.001 |
| 4o | Formatted vs I order | 100 | 0.071 [0.020, 0.121] | p=0.007 |
| 4o | Formatted vs Please | 100 | 0.071 [0.015, 0.126] | p=0.014 |
| 4o | Unformatted vs I order | 100 | -0.010 [-0.051, 0.025] | p=0.687 |
| 4o | Unformatted vs Please | 100 | -0.010 [-0.045, 0.025] | p=0.673 |
| 4o | I order vs Please | 100 | 0.000 [-0.040, 0.040] | p=1.000 |
| 4o | Formatted vs Unformatted | 90 | 0.056 [0.000, 0.111] | p=0.053 |
| 4o | Formatted vs I order | 90 | 0.061 [0.005, 0.116] | p=0.037 |
| 4o | Formatted vs Please | 90 | 0.066 [0.010, 0.126] | p=0.026 |
| 4o | Unformatted vs I order | 90 | 0.000 [-0.035, 0.035] | p=1.000 |
| 4o | Unformatted vs Please | 90 | -0.010 [-0.045, 0.025] | p=0.632 |
| 4o | I order vs Please | 90 | 0.035 [0.000, 0.071] | p=0.078 |
| 4o | Formatted vs Unformatted | 51 | 0.015 [-0.045, 0.076] | p=0.590 |
| 4o | Formatted vs I order | 51 | -0.020 [-0.081, 0.040] | p=0.575 |
| 4o | Formatted vs Please | 51 | 0.000 [-0.066, 0.066] | p=1.000 |
| 4o | Unformatted vs I order | 51 | -0.015 [-0.061, 0.030] | p=0.574 |
| 4o | Unformatted vs Please | 51 | 0.010 [-0.035, 0.051] | p=0.699 |
| 4o | I order vs Please | 51 | 0.040 [-0.000, 0.081] | p=0.065 |

**Supplementary Table 2.** Pairwise-comparison results across all conditions for each model using paired bootstrap-permutation tests (5,000 replicates) against the random baseline (25%). P-values represent the proportion of permuted differences with absolute values exceeding observed differences under the null hypothesis. Questions were considered 'perfect' at ≥100 successes, with sensitivity analyses at thresholds of 90 and 51.

| Model | Condition | Threshold | RD [95% CI] | Statistics |
|---|---|---|---|---|
| 4o mini | Formatted | 100 | 4.5% [-4.0%, 13.1%] | p=0.342 |
| 4o mini | Formatted | 90 | 7.6% [-1.5%, 16.7%] | p=0.112 |
| 4o mini | Formatted | 51 | 14.1% [4.5%, 23.2%] | p=0.003 |
| 4o mini | Unformatted | 100 | -8.1% [-15.7%, -0.5%] | p=0.039 |
| 4o mini | Unformatted | 90 | -2.0% [-10.6%, 6.1%] | p=0.638 |
| 4o mini | Unformatted | 51 | 17.2% [7.1%, 26.8%] | p=0.001 |
| 4o mini | I order | 100 | -8.6% [-16.7%, -0.5%] | p=0.032 |
| 4o mini | I order | 90 | -2.0% [-10.6%, 6.6%] | p=0.645 |
| 4o mini | I order | 51 | 12.1% [2.5%, 22.2%] | p=0.015 |
| 4o mini | Please | 100 | -9.6% [-17.7%, -2.0%] | p=0.014 |
| 4o mini | Please | 90 | -1.0% [-9.6%, 7.6%] | p=0.840 |
| 4o mini | Please | 51 | 17.7% [7.6%, 27.8%] | p<0.001 |
| 4o | Formatted | 100 | 5.1% [-3.5%, 13.1%] | p=0.263 |
| 4o | Formatted | 90 | 11.1% [2.0%, 19.7%] | p=0.015 |
| 4o | Formatted | 51 | 21.7% [12.1%, 30.8%] | p<0.001 |
| 4o | Unformatted | 100 | -3.5% [-11.6%, 4.5%] | p=0.430 |
| 4o | Unformatted | 90 | 5.1% [-4.0%, 13.6%] | p=0.284 |
| 4o | Unformatted | 51 | 20.7% [11.1%, 30.3%] | p<0.001 |
| 4o | I order | 100 | -3.0% [-11.1%, 5.1%] | p=0.443 |
| 4o | I order | 90 | 7.6% [-1.0%, 16.2%] | p=0.097 |
| 4o | I order | 51 | 23.7% [14.1%, 33.3%] | p<0.001 |
| 4o | Please | 100 | -1.5% [-9.6%, 6.6%] | p=0.716 |
| 4o | Please | 90 | 6.1% [-2.5%, 14.6%] | p=0.170 |
| 4o | Please | 51 | 20.2% [10.6%, 29.8%] | p<0.001 |

**Supplementary Table 3.** Pairwise-comparison of proportions between "Please" and "I order" conditions for GPT-4o using z-tests for two independent proportions (different tests treating them as dependent lead to similar results due to extreme differences). P-values represent the probability of observing the test statistic or more extreme values under the null hypothesis of no difference between conditions (uncorrected).

| Question # | Difference [95% CI] | Statistics |
|---|---|---|
| 158 | 0.610 [0.488, 0.702] | z=8.725, p<0.001 |
| 171 | 0.380 [0.252, 0.491] | z=5.672, p<0.001 |
| 187 | 0.353 [0.236, 0.454] | z=5.754, p<0.001 |
| 155 | 0.340 [0.215, 0.451] | z=5.198, p<0.001 |
| 2 | 0.300 [0.176, 0.412] | z=4.675, p<0.001 |
| 27 | 0.250 [0.167, 0.343] | z=5.345, p<0.001 |
| 198 | 0.250 [0.127, 0.363] | z=3.959, p<0.001 |
| 135 | 0.230 [0.095, 0.354] | z=3.327, p<0.001 |
| 57 | 0.220 [0.083, 0.346] | z=3.142, p=0.002 |
| 148 | 0.220 [0.091, 0.339] | z=3.335, p<0.001 |
| 105 | -0.633 [-0.727, -0.523] | z=-9.278, p<0.001 |
| 60 | -0.490 [-0.594, -0.362] | z=-7.088, p<0.001 |
| 10 | -0.470 [-0.578, -0.337] | z=-6.650, p<0.001 |
| 36 | -0.400 [-0.507, -0.277] | z=-6.116, p<0.001 |
| 63 | -0.400 [-0.514, -0.265] | z=-5.667, p<0.001 |
| 126 | -0.350 [-0.460, -0.226] | z=-5.375, p<0.001 |
| 3 | -0.340 [-0.455, -0.209] | z=-5.009, p<0.001 |
| 7 | -0.340 [-0.442, -0.231] | z=-5.804, p<0.001 |
| 169 | -0.290 [-0.413, -0.152] | z=-4.106, p<0.001 |
| 112 | -0.280 [-0.403, -0.142] | z=-3.967, p<0.001 |